\pdfoutput=1

\documentclass[11pt]{article}

\usepackage{acl}

\usepackage{times}
\usepackage{latexsym}

\usepackage[T1]{fontenc}

\usepackage[utf8]{inputenc}

\usepackage{microtype}
\usepackage{booktabs}
\usepackage{graphicx}

\title{Exploring Cultural Nuances in Emotion Perception Across 15 African Languages}

\author{
Ibrahim Said Ahmad$^{1,2}$, Shiran Dudy$^{1}$, 
\textbf{Tadesse Destaw Belay}$^{3,4}$, \\ \textbf{Idris Abdulmumin}$^5$, \textbf{Seid Muhie Yimam}$^{6}$, \textbf{Shamsuddeen Hassan Muhammad}$^{2,7}$, \\ \textbf{Kenneth Church}$^{1}$ \\\footnotesize{$^{1}$Northeastern University, $^{2}$Bayero University Kano, $^{3}$Instituto Politécnico Nacional, $^{4}$Wollo University,} \\ \footnotesize{$^{5}$University of Pretoria, $^6$University of Hamburg, $^7$Imperial College London} \\
\footnotesize{\texttt{\textbf{correspondence}: isahmad.it@buk.edu.ng}
}
}

\begin{document}
\maketitle

\begin{abstract}

Understanding how emotions are expressed across languages is vital for building culturally-aware and inclusive NLP systems. However, emotion expression in African languages is understudied, limiting the development of effective emotion detection tools in these languages. In this work, we present a cross-linguistic analysis of emotion expression in 15 African languages. We examine four key dimensions of emotion representation: text length, sentiment polarity, emotion co-occurrence, and intensity variations. Our findings reveal diverse language-specific patterns in emotional expression---with Somali texts typically longer, while others like IsiZulu and Algerian Arabic show more concise emotional expression. We observe a higher prevalence of negative sentiment in several Nigerian languages compared to lower negativity in languages like IsiXhosa. 
Further, emotion co-occurrence analysis demonstrates strong cross-linguistic associations between specific emotion pairs (anger-disgust, sadness-fear), suggesting universal psychological connections. Intensity distributions show multimodal patterns with significant variations between language families; Bantu languages display similar yet distinct profiles, while Afroasiatic languages and Nigerian Pidgin demonstrate wider intensity ranges. These findings highlight the need for language-specific approaches to emotion detection while identifying opportunities for transfer learning across related languages.

\end{abstract}

\section{Introduction}
The expression and perception of emotions are fundamental aspects of human communication, deeply intertwined with language and culture \cite{mesquita1997culture,wierzbicka1999emotions}. As natural language processing (NLP) advances, understanding how emotions are conveyed across different languages has become increasingly important. This paper explores the intricate relationship between emotion intensity, co-occurrence, and textual expression in various languages, aiming to uncover both universal trends and language-specific nuances.

Emotions play a crucial role in shaping interpersonal interactions, influencing not only how we communicate but also how we interpret messages \cite{jozefien2011}. Previous research has primarily focused on emotion recognition and sentiment analysis within single or multiple languages, often overlooking the rich diversity of emotional expression across linguistic boundaries \cite{tatariya-etal-2024-sociolinguistically, saravia-etal-2018-carer, havaldar-etal-2023-multilingual, demszky-etal-2020-goemotions}. Our study seeks to bridge this gap by examining emotion intensity and co-occurrence patterns in a multilingual context, thereby contributing to a more holistic understanding of emotion in language.

In this paper, we address the following research questions:

\begin{enumerate}
    \item  Is there a correlation between the length of text and the intensity or type of emotions expressed? Do certain languages tend to use longer or shorter texts to convey specific emotions?
    \item  How do sentiment polarity distributions vary across typologically diverse languages in social media text, and what factors contribute to the observed differences in negative sentiment prevalence?
    \item  What are the prevalent patterns of emotion co-occurrence in various languages, and how do these patterns reflect cultural norms or communication styles?
    \item  How does the average intensity of emotions such as joy, sadness, anger, and surprise vary across different languages? What cultural or linguistic factors might contribute to these differences?
\end{enumerate}

To answer these questions, we analyze a diverse dataset comprising text samples from fifteen (15) African languages, employing NLP techniques to measure emotion intensity and co-occurrence. Our findings reveal insights into the interplay between language, culture, and emotion, highlighting both shared and distinct characteristics across languages.

Our research contributes to the broader field of computational linguistics and offers practical implications for multilingual NLP applications, such as sentiment analysis, machine translation, and cross-cultural communication tools. Furthermore, our study underscores the importance of considering cultural and linguistic diversity in the development of emotion-aware technologies.


\label{sec:related}

\section{Related works}
Language and culture are deeply intertwined, a relationship extensively studied in psycholinguistics \cite{zhou2025culturetriviasocioculturaltheory}. Recent advancements in NLP have heightened the need to understand this intersection in social media and online texts, ensuring AI systems accurately represent cultural perspectives.  

Current research evaluates how LLMs reflect cultural diversity in language generation \cite{ahmad-etal-2024-generative,liu-etal-2024-multilingual, koto_tacl_a_00726, liu-etal-2024-multilingual,dudy2024ACII,cao-etal-2023-assessing}. Other efforts focus on building resources for culturally aware systems, such as CultureBank, a knowledge base for culture-specific data \cite{shi-etal-2024-culturebank}.  Emotion datasets across cultures, such as \citet{muhammad2025brighterbridginggaphumanannotated}, \citet{belay-etal-2025-evaluating} and \citet{demszky-etal-2020-goemotions}, further enable training inclusive AI models. Recent work also adopts multimodal approaches, analyzing cultural emotions in both text and images \cite{mohamed-etal-2024-culture, khanuja-etal-2024-image}.  

While existing works leverage social media data to study cultural emotions, comparative analyses across languages remain limited. For instance, studies often focus on developing datasets rather than systematically contrasting how emotions manifest in different linguistic contexts. 

Social media’s informal, code-switched, and multilingual nature offers a unique opportunity to investigate these dynamics, yet few works explicitly compare cross-cultural emotional patterns between languages, such as Arabic versus Hausa or Amharic versus Igbo. Bridging this gap could uncover nuanced cultural divergences and improve AI’s adaptability to global linguistic diversity.


\section{Experiments}

\subsection{Data}
Our study utilizes existing datasets by \citet{muhammad2025brighterbridginggaphumanannotated,muhammad-etal-2025-semeval} and \citet{belay-etal-2025-evaluating} that encompass text samples from multiple languages. We focus on 15 African languages: 
Algerian Arabic (ary), Amharic (amh), Emakhuwa (vmw), Hausa (hau), Igbo (igo), IsiXhosa (xho), IsiZulu (zul), Kinyarwanda (kin), Mozambique Portuguese (ptmz), Nigerian Pidgin (pcm), Oromo (orm), Swahili (swa), Somali (som), Tigrinya (tir), and Yoruba (yor).

The dataset consists of a collection of texts from two sources, social media posts and news articles headlines, ensuring a comprehensive representation of emotional expression in different contexts.

\subsection{Methodology}

As the goal of this project was to study the relation of emotional expression across language and culture, in the next sections, we provide a description of how we assess these relations. Our analysis focused on the following questions: Are there language differences in emotional expression? Can we characterize languages (and their specific sources) by their length? is emotion co-occurrence unique or similar across languages? 
We analyzed different key dimensions of emotional expression, such as text length distribution, emotion co-occurrence in the texts, and emotion intensity analysis. 


\section{Results}

\subsection{Text Length Analysis}

\begin{figure*}[h]
\centering
\includegraphics[width=0.9\linewidth]{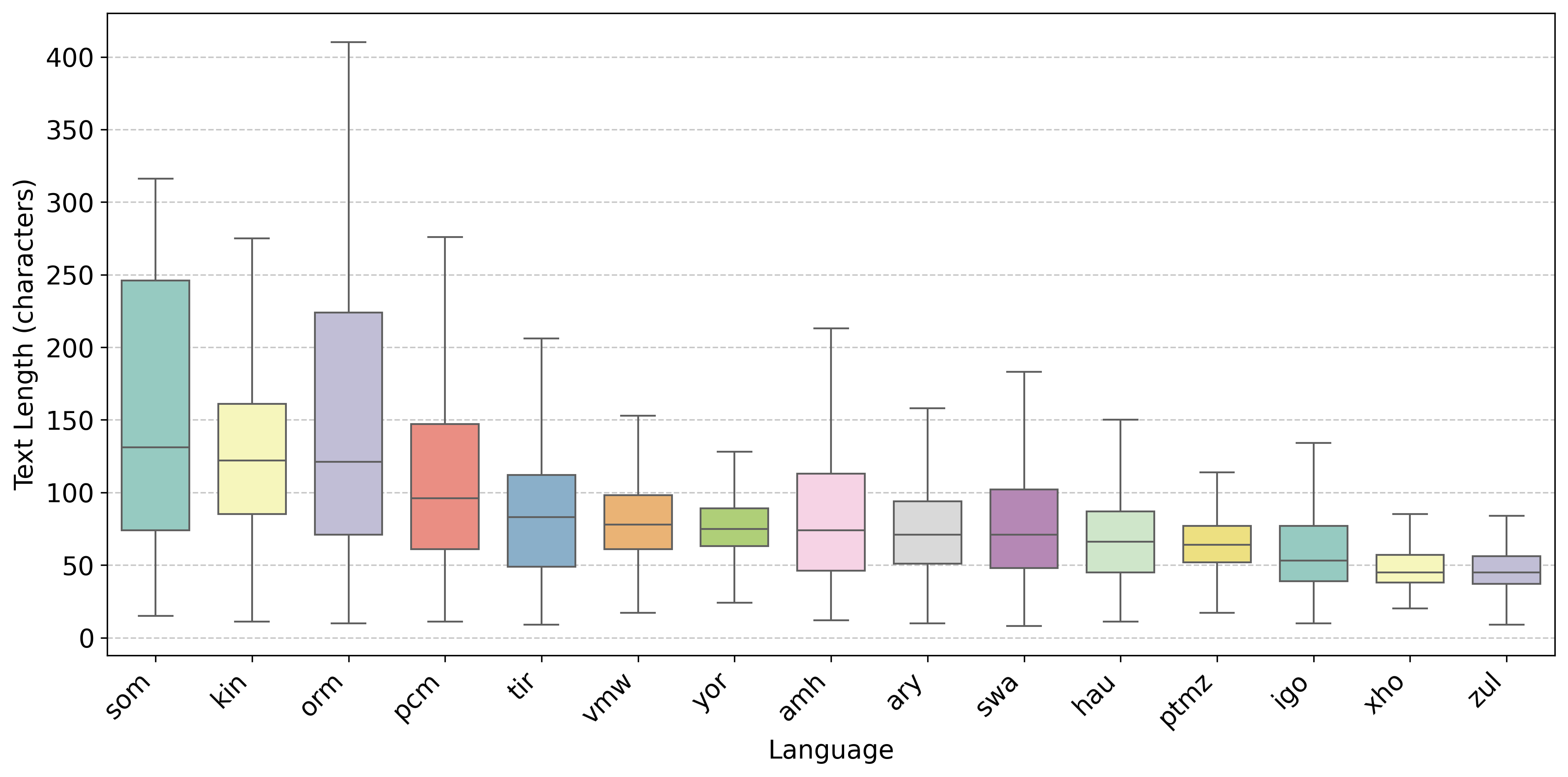}
\caption{ Text Length Distribution by Language and Emotion with Language Comparison.
This box plot displays the distribution of text length (in characters) for different emotions across all the languages, sorted by median text length. Languages like Somali,  Kinyarwanda and Oromo have significantly longer text expressions across most emotions, with some texts exceeding 300 characters. Anger and sadness generally have longer text expressions across most languages. As languages move to the right of the chart, text length decreases substantially, with IsiXhosa and IsiZulu having the shortest text expressions regardless of emotion type.
}
\label{fig:text-length}
\end{figure*}

Figure~\ref{fig:text-length} shows the distribution of the length of text in various languages. Notably, there is a significant variation in the median text lengths among languages. For example, languages such as Somali (som) exhibit a higher median text length compared to others like Algerian Arabic (ary) and IsiZulu (zul), which have relatively shorter median lengths. This disparity suggests that certain languages may inherently require more characters to convey meaning, or it could reflect cultural or linguistic differences in communication styles.

Additionally, the spread of the interquartile range (IQR) and the presence of outliers indicate that some languages have more consistent text lengths, while others show greater variability. Languages with wider IQRs and more outliers, such as Nigerian Pidgin (pcm) and Oromo (orm), may be used in more diverse contexts or have more flexible grammatical structures that allow for varied text lengths. This is not surprising for Nigerian Pidgin, as it is a variant of English mixed with several local languages \cite{Akande+2010+3+22}. 

\begin{figure*}[h]
\centering
\includegraphics[width=1.0\linewidth]{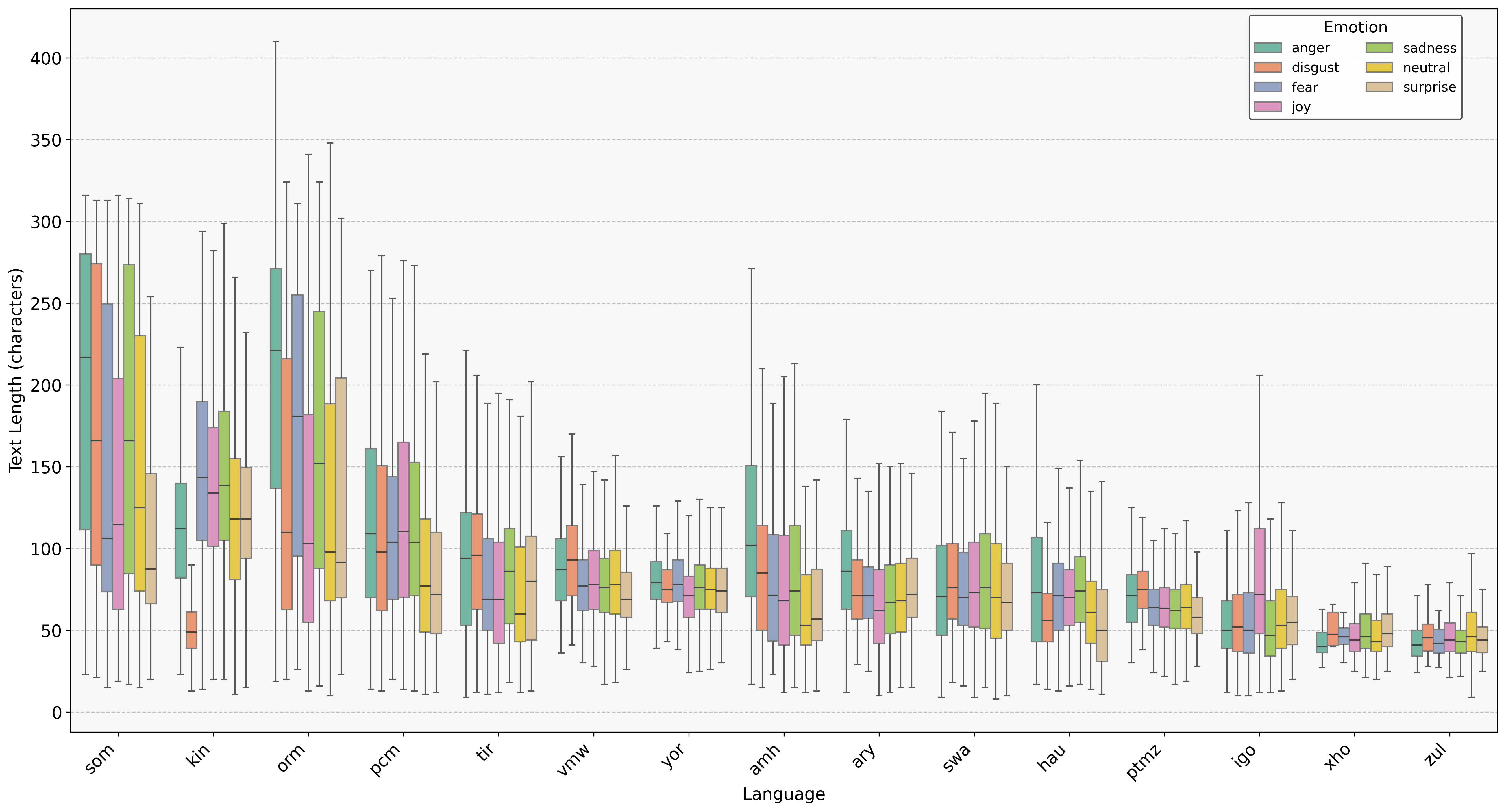}
\caption{Text Length Distribution by Language Regardless of Emotion.
This box plot shows the overall text length distribution for each language, sorted by median length. Somali has the longest texts with a median of approximately 130 characters and maximum values exceeding 300 characters. There's a clear gradient with Kinyarwanda and Oromo also showing relatively long text expressions, while languages on the right side have consistently short expressions with medians below 50 characters. This indicates significant variation in typical text length across different languages.}
\label{fig:cultural-expression}
\end{figure*}

Figure~\ref{fig:cultural-expression} extends the analysis by incorporating emotional context. Each language is further categorized by the lengths of texts associated with particular emotions, revealing how different emotions influence text length within each language. The emotions analyzed are sadness, joy, fear, neutral, anger, surprise, and disgust.

Overall, the text length associated with different emotions within a dataset tends to show similar median scores and confidence intervals. This indicates that, aside from a few outliers, emotions may not be easily distinguishable if the source is known based on their length distribution. However, since these distributions exhibit consistent patterns within the same source, a given distribution could potentially reveal the dataset or source from which they originated (yet not its emotional label).



These findings have several implications for linguistic research, natural language processing, and cross-cultural communication studies. Understanding the relationship between text length, language, and emotion can inform the development of more nuanced language models and improve the accuracy of sentiment analysis tools. Additionally, these insights can aid in creating more effective communication strategies in multilingual and multicultural settings.

\subsection{Sentiment Polarity Distribution}
Table ~\ref{tab:sentiments} shows the sentiment polarity across various languages, which reveals how sentiments are expressed differently in each language. The emotions were categorized into positive, negative, and neutral sentiments, to highlights key patterns in emotional expression. Emotions joy and surprise were considered as positive, while anger, fear, and disgust were categorized as negative. Understanding sentiment distributions can inform the development of more accurate sentiment analysis tools and help tailor messages to align with the emotional expectations of different linguistic groups.

Neutral sentiments dominate across most languages, particularly in IsiZulu (zul), Yoruba (yor), Emakhuwa (vmw), and Mozambique Portuguese (ptmz). This does not necessarily suggest that a significant portion of communication in these languages is factual or informative, but could likely be as a result of part of the data being collected from news headlines that contain more neutral emotions. However, what is very surprising is the variation between IsiZulu (zul) and IsiXhosa (xho) as these languages are very similar, and collected from the same source.  

The distribution of positive and negative sentiments varies significantly. A striking observation is that all Nigerian languages, except for Yoruba (yo), exhibit significantly higher negative sentiment. Previous studies have documented the prevalence of hateful and toxic content in the Nigerian social media space \cite{jimada2023social, muhammad2025afrihatemultilingualcollectionhate}.

For most of the other languages, there is a unique sentiment pattern. For example, Algerian Arabic (ary) has a high count of both negative and positive sentiments, suggesting a diverse emotional range. Meanwhile, IsiXhosa (xho) has only 9.1\% negative sentiment, which contrasts with the typical characteristics of social media texts and requires further investigation.


\begin{table}[htbp]
\centering
\caption{Cross-Linguistic Sentiment Polarity Distribution with Neutral dominance in IsiZulu (zul) and Yoruba (yor) contrasts with varied Positive/Negative expressions in Hausa, Swahili, and others, offering insights for culturally-tuned sentiment analysis and communication strategies}
\begin{tabular}{lrrr}
\toprule
\textbf{Language} & \textbf{Neg(\%) $\downarrow$} & \textbf{Pos(\%)} & \textbf{Neu(\%)} \\
  \midrule
  \textbf{\texttt{pcm}} & 55.3 & 37.2 & 7.5 \\
  \textbf{\texttt{ibo}} & 51.3 & 22.4 & 26.3 \\
  \textbf{\texttt{tir}} & 49.8 & 21.3 & 28.9 \\
  \textbf{\texttt{hau}} & 46.8 & 35.7 & 17.5 \\
  \textbf{\texttt{amh}} & 45.9 & 24.4 & 29.7 \\
  \textbf{\texttt{orm}} & 35.0 & 37.8 & 27.3 \\
  \textbf{\texttt{kin}} & 34.6 & 26.8 & 38.6 \\
  \textbf{\texttt{ary}} & 30.3 & 36.1 & 33.6 \\
  \textbf{\texttt{som}} & 29.3 & 25.7 & 45.0 \\
  \textbf{\texttt{ptmz}} & 20.2 & 23.5 & 56.2 \\
  \textbf{\texttt{vmw}} & 20.2 & 23.5 & 56.3 \\
  \textbf{\texttt{swa}} & 18.7 & 30.8 & 50.5 \\
  \textbf{\texttt{yor}} & 13.4 & 21.2 & 65.4 \\
  \textbf{\texttt{zul}} & 12.5 & 15.3 & 72.3 \\
  \textbf{\texttt{xho}} & 9.1 & 66.2 & 24.6 \\
\hline
\end{tabular}
\label{tab:sentiments}
\end{table}

\subsection{Emotion Co-occurrence Patterns}

\begin{figure*}[h]
\centering
\includegraphics[width=0.6\linewidth]{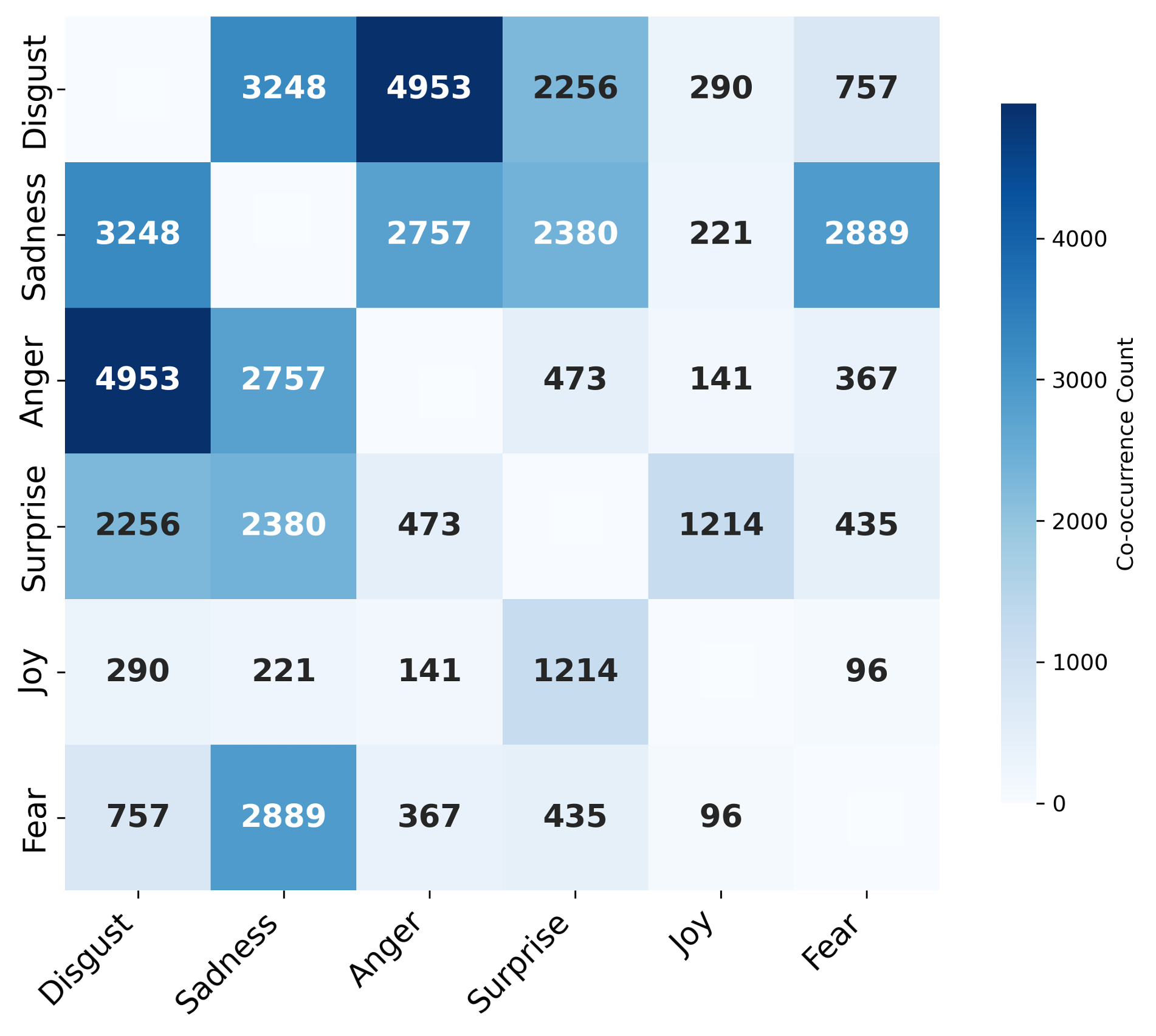}
\caption{Emotion Co-occurrence Across All Languages. Negative emotions such as Disgust, Sadness, and Anger frequently co-occur, with Disgust and Anger having the highest co-occurrence. In contrast, Joy has the lowest co-occurrence with other emotions, particularly with Disgust and Anger. Sadness and Fear also frequently co-occur, indicating a strong relationship between these emotions in the dataset.}
\label{fig:co-occurrence}
\end{figure*}

Figure~\ref{fig:co-occurrence} presents a heatmap visualization of the frequency with which different emotions co-occur across all languages. This analysis is crucial as it provides insights into the interplay of emotions expressed in human language across diverse linguistic contexts.  

Sadness exhibits the highest co-occurrence frequency, particularly with other negative emotions. The strongest co-occurrence is observed between anger and disgust (4953 occurrences), followed by sadness and disgust (3248 occurrences) and sadness and fear (2889 occurrences). This aligns with previous findings on social media data, where negative emotions tend to reinforce each other, often within discussions of distressing or controversial topics.  

Joy, on the other hand, shows the lowest co-occurrence with negative emotions, suggesting a clearer separation between positive and negative sentiment expressions in written text. However, joy and surprise (1214 occurrences) co-occur more frequently than joy with any negative emotion, reinforcing the idea that surprise can function as both a positive and negative emotional response, depending on context.  

Surprise exhibits moderate co-occurrence with both positive and negative emotions, reflecting its contextual flexibility. It can be associated with shock and fear in negative contexts or delight in positive ones. Notably, surprise and sadness (2380 occurrences) co-occur more frequently than surprise and joy, hinting at a potential bias in social media conversations where unexpected events are more often framed negatively.  

A particularly significant finding is the strong co-occurrence of fear and sadness across languages (2889 occurrences), suggesting a fundamental psychological connection between these emotions that transcends cultural boundaries. Fear is often linked to anxiety, while sadness is associated with depression, and many studies in mental health have documented the frequent co-occurrence of depression and anxiety \cite{Malygin2022-as, 10.1001/jamapsychiatry.2020.0601}. This highlights the importance of considering emotional interdependencies in sentiment analysis and mental health studies.  

Additionally, some co-occurrence asymmetries can be observed, where the pairing of two emotions does not always exhibit equal frequency in both directions. This could be influenced by linguistic variations in how emotions are expressed or by structural differences in text, such as whether emotions appear together in phrases or in separate clauses within the same sentence.

\subsection{Emotion Intensity Distribution}

\begin{figure*}[h]
\centering
\includegraphics[width=0.9\linewidth]{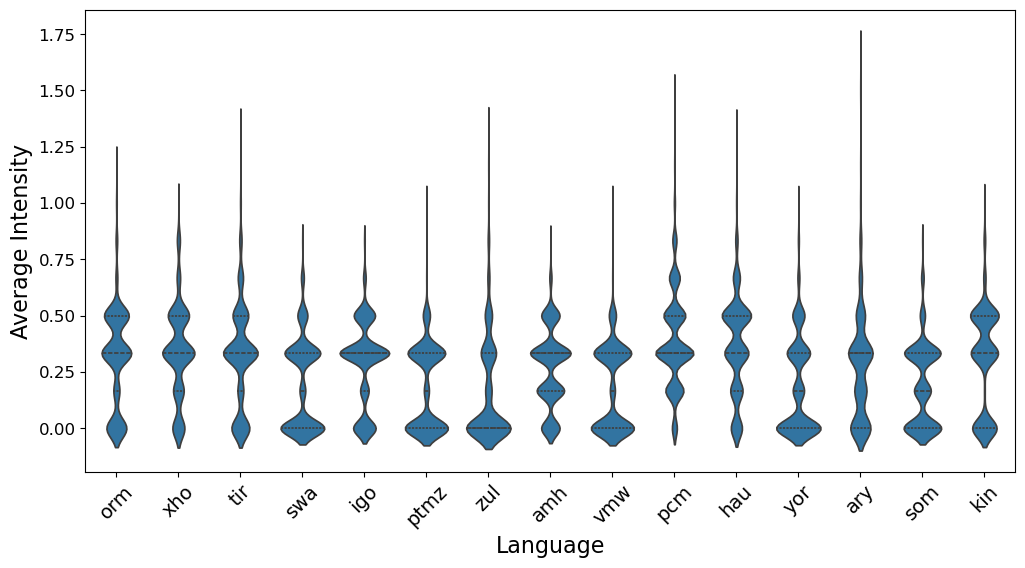}
\caption{Emotion Intensity Distribution by Language Using Violin Plots.
This violin plot depicts the distribution of emotion intensity scores across the languages. The width of each "violin" shows the density of data points at that intensity level. Languages like Tigrinya (tir)  and Algerian Arabic (ary) show the greatest variation in emotion intensity, with distributions extending to higher values. Most languages show a concentration of intensity values around 0.25-0.5, indicating that moderate emotion intensity is most common across languages. Some languages like IsiZulu (zul) and Swahili (swa) have narrower distributions, suggesting more consistent emotion intensity levels.}
\label{fig:intensity-distribution}
\end{figure*}

Figure~\ref{fig:intensity-distribution} presents the emotion intensity distributions across languages. A violin plot was used to provide a detailed visualization of the data distribution beyond traditional box plots, capturing the density and spread of emotion intensities.  

The plot reveals significant patterns in emotion intensity across languages, offering insights into how emotions are conveyed in written text across typologically diverse languages.  

The intensity distributions demonstrate both cross-linguistic patterns and language-specific characteristics. Most languages exhibit multimodal distributions with primary concentrations in the 0.25–0.5 range, suggesting some universal tendencies in how emotions are expressed in text. However, notable variations emerge when examining specific language families.  

Bantu languages (zul, xho, swa, vmw, kin) generally display similar distribution patterns, though with meaningful differences. While IsiZulu (zul) shows a narrower distribution primarily centered below 0.5, closely related IsiXhosa (xho) exhibits slightly higher variability with moderate outliers. This suggests subtle distinctions in how emotions are articulated in text, even between genetically related languages.  

Afroasiatic languages in the dataset (orm, tir, amh, som, ary) display considerable internal variation. Tigrinya (tir) and Oromo (orm) have higher outlier values reaching approximately 1.4 and 1.25, respectively, while Somali (som) presents a distinct bimodal distribution with peaks near 0 and 0.3. This may reflect linguistic structures influencing written emotional expression, such as sentence structure and common lexical choices.  

The most pronounced variation appears in Nigerian Pidgin (pcm) and Hausa (hau), which demonstrate wider distributions with higher intensity values. This could reflect the influence of expressive language use, code-switching, or the diversity of linguistic influences shaping emotional expression in these languages.  

These distribution patterns have significant implications for multilingual NLP, particularly in emotion-aware text processing. The consistent multimodal patterns suggest the potential for transfer learning approaches, while the language-specific variations highlight the need for tailored models that account for different ways emotions are expressed in text.  

The bimodal distributions observed in several languages (swa, ptmz, vmw, som) warrant particular attention, as they may require specialized approaches to modeling emotional intensity in text classification and sentiment analysis tasks. These findings underscore the importance of language-specific modeling in multilingual NLP applications.

\section{Discussion}
\begin{enumerate}  
    \item \textbf{Text Length and Emotion Expression:}  
    The analysis revealed significant cross-linguistic variation in text length, with languages like Somali requiring longer texts compared to Algerian Arabic or IsiZulu. This may reflect inherent linguistic structures (e.g., agglutination) or cultural communication styles favoring brevity or elaboration. While emotions showed minimal differentiation in text length within languages, the consistent patterns observed within datasets suggest that text length distributions could serve as indirect identifiers of data sources. This finding emphasizess the need for language-specific preprocessing in NLP tasks to account for structural differences. Future work should explore whether text length differences persist in controlled contexts (e.g., translations of the same content) to disentangle linguistic and cultural influences.  

    \item \textbf{Sentiment Polarity Distributions:}  
    The dominance of neutral sentiments in languages like IsiZulu and Yoruba likely stems from data source biases (e.g., news headlines), emphasizing the need for diversified corpora in sentiment analysis. The clear contrast between similar languages (IsiZulu vs. IsiXhosa) challenges assumptions about linguistic homogeneity in emotional expression and warrants investigation into dialectal or sociocultural factors. The high prevalence of negative sentiment in Nigerian languages aligns with documented toxic content trends in Nigerian social media, suggesting that region-specific sociopolitical contexts heavily influence sentiment patterns. These results highlight the importance of contextualizing sentiment analysis tools to account for both linguistic features and cultural realities.  

    \item \textbf{Emotion Co-occurrence Patterns:}  
    The frequent co-occurrence of negative emotions (e.g., anger-disgust, sadness-fear) reflects their psychological interconnectedness and the tendency for distressing topics to evoke multiple negative responses in text. Joy’s separation from negative emotions supports its role as a distinct affective category, while surprise’s association with both positive and negative contexts highlights its contextual dependence. The cross-cultural consistency in fear-sadness co-occurrence aligns with mental health research linking anxiety and depression, suggesting that emotional interdependencies in text may mirror real-world psychological phenomena. Asymmetries in co-occurrence frequencies (e.g., surprise-sadness vs. surprise-joy) could inform culturally nuanced NLP models by prioritizing emotion pairs prevalent in specific linguistic contexts.  

    \item \textbf{Emotion Intensity Variation:}  
    Bantu languages’ shared intensity patterns contrast with Afroasiatic languages’ internal variability, indicating that language family typology influences emotional expression. The bimodal distributions in Somali and Nigerian Pidgin may arise from code-switching or source-specific text conventions. Higher intensity outliers in Hausa and Tigrinya suggest these languages employ more emphatic lexical or syntactic strategies for emotional emphasis.   
\end{enumerate}




\subsection{Limitations and Future Work}
\begin{enumerate}
   
 \item This study relies on existing datasets of social media (X) posts and news headlines across fifteen African languages, which presents several important constraints. These data sources were not originally designed for cross-linguistic emotion analysis and may contain uneven representation across languages. Additionally, social media content and headlines represent specialized forms of communication characterized by brevity, informality, and often heightened emotional states that may not reflect the full spectrum of emotional expression in these languages.

\item Since this study spans multiple languages, future work could explore whether specific linguistic or cultural factors influence these co-occurrence patterns. For example, do tonal languages like Hausa or Yoruba exhibit different co-occurrence tendencies compared to non-tonal languages? Such an analysis would provide deeper insights into how language structure affects emotional expression.  

\item Our text-based approach cannot capture non-verbal aspects of emotional communication such as facial expressions. The brevity of tweets and headlines also limits our ability to analyze complex emotional narratives or subtle transitions that emerge in longer discourse. Future work would benefit from purpose-built datasets with diverse text genres and multimodal data to provide a more comprehensive understanding of emotional expression across languages.


\end{enumerate}
\section{Conclusion}
In this paper, we present a comprehensive cross-linguistic analysis of emotional expression in fifteen different languages. Our findings reveal significant variations in how emotions are expressed textually, with distinct patterns in text length, emotion co-occurrence, intensity, and distribution. These variations highlight the importance of considering linguistic and cultural factors in computational approaches to emotion analysis.

The observed differences challenge universal models of emotional expression and underscore the need for language-specific approaches in multilingual NLP systems. By recognizing and accounting for these cross-linguistic variations, we can develop more culturally sensitive and accurate computational models for emotion detection and analysis.

Our work contributes to both theoretical understanding of cross-cultural emotional expression and practical implementation of multilingual emotion recognition systems. Future research should continue to explore the interplay between language, culture, and emotional expression across a broader range of languages and communicative contexts.

\bibliography{anthology,custom}
\bibliographystyle{acl_natbib}

\end{document}